\documentclass[conference]{IEEEtran}

\usepackage{graphicx} % For figures
\usepackage{amsmath, amssymb} % For math symbols and environments
\usepackage{cite} % For better citation handling
\usepackage{url}

\begin{document}

\title{Object Location Prediction in Real-time using LSTM Neural Network and Polynomial Regression}

\author{%
\IEEEauthorblockN{Petar Stojković}
\IEEEauthorblockA{Email: petar.stojkovic998@outlook.com}
\and
\IEEEauthorblockN{Predrag Tadić}
\IEEEauthorblockA{Department of Signals and Systems\\
University of Belgrade School of Electrical Engineering\\
Email: tadicp@etf.bg.ac.rs}
}

\maketitle

\begin{abstract}
This paper details the design and implementation of a system for predicting and interpolating object location coordinates. Our solution is based on processing inertial measurements and global positioning system data through a Long Short-Term Memory (LSTM) neural network and polynomial regression. LSTM is a type of recurrent neural network (RNN) particularly suited for processing data sequences and avoiding the long-term dependency problem. We employed data from real-world vehicles and the global positioning system (GPS) sensors. A critical pre-processing step was developed to address varying sensor frequencies and inconsistent GPS time steps and dropouts. The LSTM-based system's performance was compared with the Kalman Filter. The system was tuned to work in real-time with low latency and high precision. We tested our system on roads under various driving conditions, including acceleration, turns, deceleration, and straight paths. We tested our proposed solution's accuracy and inference time and showed that it could perform in real-time. Our LSTM-based system yielded an average error of 0.11 meters with an inference time of 2 ms. This represents a 76\% reduction in error compared to the traditional Kalman filter method, which has an average error of 0.46 meters with a similar inference time to the LSTM-based system.
\end{abstract}


\section{Introduction}
Global Positioning System (GPS) technology has revolutionized our ability to navigate, track, and monitor locations. As its applicability expands, especially in sectors like transportation, logistics, and emergency response, the demand for accurate real-time GPS tracking and prediction has surged. Traditional methods, such as linear regression and Kalman Filters, possess inherent limitations in addressing this challenge. This paper introduces a novel approach combining Long Short-Term Memory (LSTM) neural networks, designed to model sequential data, with polynomial regression to predict GPS coordinates in real-time.

LSTM, a type of Recurrent Neural Network (RNN), overcomes the vanishing gradient problem plaguing traditional RNNs as described in \cite{feng2017}. Its architecture employs memory cells and gates to control information flow, making it optimal for tasks such as time series forecasting. This work leverages LSTM's ability to predict sequential steps based on prior actions, integrating it with polynomial regression to capture data trends with real-time vehicle timestamps.

The subsequent sections detail the system's design, research methodology, results, and discussions on its potential and limitations. 

\begin{figure}
    \centering
    \includegraphics[width=\linewidth]{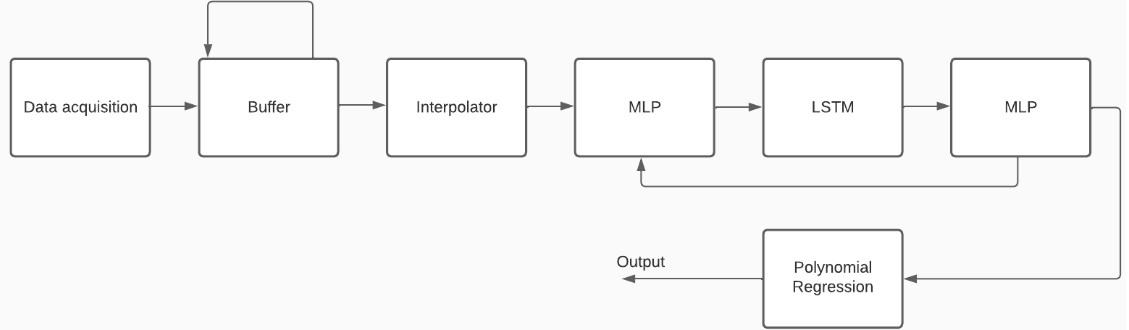}
    \caption{Block diagram of system from predicting current location} % Optional, but it's good to have
    \label{fig:block} % For referencing the figure later in the text (if needed)
\end{figure}

\section{Related Work}

The primary emphasis of this work lies in its ability to operate in real-time with very limited resources, a critical factor often not addressed by other works. While various studies have delved into the prediction of vehicle trajectories \cite{yang2020, altche2023} or the correction of GPS sensor errors \cite{liu2023}, our system uniquely focuses on predicting not only the future trajectory but also the exact current position of a vehicle. Specifically, the model fills the gap between the last recorded GPS sample and a prediction that stands 0.2 seconds ahead, accounting for scenarios with low data sampling frequencies or delayed updates. This continuous real-time prediction ensures that the system can seamlessly adapt to fluctuating data streams and guarantee accurate position tracking even in challenging circumstances. A key highlight is the system's efficiency in operation speed and minimal resource consumption. An innovative aspect of this work is its capability to take control when there is data dropout from the GPS sensor (either due to outliers or noise). 

For accuracy and security, the model leverages data from both GPS and the inertial measurement unit (IMU) sensors. In case of GPS data dropouts, the model employs its own predictions in place of missing observations until valid GPS data becomes available, ensuring that the system always has access to the vehicle's current position and operates stably. 

Several works, while not identical in intent, exhibit similarities or offer insights into the methods applied in this study. Specifically, papers \cite{yang2020} and \cite{gao2018} drew significant attention:

\begin{enumerate}
    \item In \cite{yang2020}, the authors focus on predicting a ship's position, speed, and orientation using sizable sequences of AIS (Automatic Identification System) data as inputs, which include various parameters like Speed Over Ground, Universal Time Coordinated, Course Over Ground, Latitude, and Longitude. A straightforward LSTM network, complemented with normalization techniques, is applied to make these predictions. 
    Drawing parallels with our study, there is a shared idea of the LSTM network for predicting position coordinates. However, it's crucial to note that the results of their approach are not readily aligned with ours due to differences in the application domain and the nature of the data processed. Their model emphasizes maritime navigation, processing an array of AIS data, whereas our model is based on vehicular movement, using data from GPS and IMU sensors to make real-time predictions on roads under various driving conditions. Despite these distinctions, the conceptual resemblance in employing LSTM networks for predictive analysis in navigational contexts is noteworthy.
    
    \item The research presented in \cite{liu2023} exhibits a conceptual resemblance to our work, focusing on enhancing GPS coordinate accuracy using LSTM networks. Their approach concentrates exclusively on utilizing GPS data, incorporating a smoothing technique to correct errors influenced by various factors like satellite clock discrepancies, atmospheric delays, and multipath fading. However, our study extends beyond this scope by not only targeting the prediction of future trajectories but also focusing on the real-time accuracy of an object’s current position. We synchronize data from both GPS and inertial measurement units (IMUs), and our model is robust enough to maintain operational stability even during GPS data dropouts, ensuring uninterrupted precision in object position estimation. While there are parallels in employing LSTM networks, methodologies, and objectives, it is hard to compare the results because of the different goals of the projects.
    
    \item The research conducted by Altché in \cite{altche2023} bears some resemblance to our work, as it delves into predicting vehicle trajectories on highways by employing GPS data and lane descriptions. Their model dominantly focuses on understanding the intentions of surrounding vehicles, using a long short-term memory (LSTM) neural network to predict future longitudinal and lateral trajectories for vehicles on the highway. However, it's crucial to highlight that their modeling concept, while sharing superficial similarities with ours, diverges significantly in terms of objectives and application. Altché’s work is more centered around enhancing the safety and efficiency of autonomous vehicles by improving medium-term forecasts in various traffic densities, helping in making more informed decisions based on the predicted trajectories of surrounding vehicles.
    
    In contrast, our study is meticulously engineered towards predicting and interpolating object location coordinates in real-time with heightened precision and reduced latency. We navigate through the challenges posed by varying sensor frequencies and inconsistent GPS time steps to ensure the accuracy and reliability of our predictions under various driving conditions. While both studies employ LSTM neural networks, the distinction lies in the application, with our work striving for enhanced accuracy in real-time GPS tracking and prediction, ensuring seamless adaptation to fluctuating data streams and maintaining operational stability even under challenging circumstances.
        
    \item The research in \cite{gao2018} utilized a compelling model, BI-LSTM-RNN. The bidirectional component improves the model by allowing it to access future contexts or inputs, not just the past ones, making the predictions more correlated or consistent with each other. This architecture helps in preventing the model from memorizing the training data (overfitting) due to repetitive patterns (autocorrelation) present in the data, making the model more generalized and capable of handling unseen data.
    While there is similarity with our work in employing LSTM layers, the bidirectional architecture of the model in \cite{gao2018} creates a divergent pathway from ours, focusing on a more intertwined relationship between future and past events in predicting coordinates. It’s essential to underline that while both studies aim at enhancing the accuracy and reliability of positional predictions, he modeling concepts are fundamentally different, which makes the results interesting but hard to compare directly.
    
\end{enumerate}
\textbf{Note:} The most significant difference between all mentioned papers and ours is that our approach is focused on real-time optimization with additions of the current to the future positions of the vehicle.

\section{Methodology}

\subsection{System Overview}

The main goal of our proposed system is to provide short-term predictions of the vehicle's position. Based on acceleration and yaw angle data from the vehicle's inertial measurement unit (IMU), and on previous GPS data observed up to current time $t$, we predict the future position in two steps. First, we use an LSTM to get a single predicted position 200 milliseconds into the future. Then, we use a polynomial regression model to obtain estimates on a more fine-grained time scale between $t$ and $t + 200\, \mathrm{ms}$.   

Figure \ref{fig:block} illustrates the complete prediction process. The key steps are as follows.
\begin{enumerate}
    \item IMU and GPS data are sampled asynchronously and with very different sampling intervals. In order to facilitate the task of the LSTM, we interpolate the data to obtain synchronous samples, which are equidistant in time.  
    \item The preprocessed data form a sequence that is fed to an LSTM which extrapolates the sequence, thus forecasting the next data point.
    \item Our experiments showed that better performance is obtained by letting the LSTM operate on a different, learned representation of the input data, rather then on the raw inputs themselves. These representation, or embeddings, are computed by a multi-layer perceptron (MLP), which preceeds the LSTM block. Likewise, the output of the LSTM is also processed by a separate MLP, to map the predicted representations back into predictions of the raw input data.
    \item The LSTM block computes just a single prediction for the vehicle's position $200\,\mathrm{ms}$ into the future. To obtain estimates of position at arbitrary times within the $200\,\mathrm{ms}$ interval between the last observation and the prediction, we again employ a polynomial regression model.
\end{enumerate}

\subsection{Data Collection and Preprocessing}

One of the core challenges in vehicle movement tracking arises from the myriad of data sources available, ranging from Lidar sensors and cameras to simpler measurements such as GPS coordinates, travelled distance, and steering metrics. In this project, we primarily harness data from GPS coordinates coupled with orientation and acceleration. Specifically, the IMU sensor, used for tracking acceleration and angular velocity, offers insights into acceleration in both the x and y axes as well as orientation metrics such as pitch and yaw angles.

GPS sensors provide data in the form of latitude and longitude, which, for the purposes of this research, can be conveniently translated into Cartesian coordinates ($x$ and $y$ axes). Such a transformation often hinges on selecting the right map projection technique, with options spanning from Mercator to Lambert projections. One of the most prevalent choices, particularly suited for GPS data, is the Universal Transverse Mercator (UTM) projection. This method translates the Earth's surface onto a 2D plane, upholding the integrity of local angles and shapes, and, in the process, allocating a unique set of x and y coordinates to every location. The conversion from GPS to UTM coordinates is contingent on several parameters, including the ellipsoidal model of Earth's shape, the central meridian of the UTM zone, and the origin of the UTM system.

To offer a clearer picture, below is an illustrative formula to convert GPS coordinates to their corresponding UTM coordinates taken from \cite{mapping}:

1. Calculation of the Meridional Arc
\begin{multline}
M = a (1 - e^2) \Bigg( \frac{1 - \frac{e^2}{4} - \frac{3e^4}{64} - \frac{5e^6}{256}}{(1 - e^2)} \phi - \\
\frac{3(1 - e^2)\left(\frac{3e^2}{8} + \frac{3e^4}{32} + \frac{45e^6}{1024}\right)}{2} \sin 2\phi + \\
\frac{15(1 - e^2)\left(\frac{15e^4}{256} + \frac{45e^6}{1024}\right)}{16} \sin 4\phi - \\
\frac{35(1 - e^2) \frac{315e^6}{3072}}{48} \sin 6\phi \Bigg)
\end{multline}

2. Calculation of x (Easting)
\begin{multline}
x = x_0 + k_0 \nu \Bigg( A + \frac{(1 - T + C) A^3}{6} + \\
\frac{(5 - 18T + T^2 + 72C - 58e'^2) A^5}{120} \Bigg)
\end{multline}

3. Calculation of y (Northing)
\begin{multline}
y = y_0 + k_0 \Bigg( M + \nu \tan \phi \Bigg( \frac{A^2}{2} + \\
\frac{(5 - T + 9C + 4C^2) A^4}{24} + \\
\frac{(61 - 58T + T^2 + 600C - 330e'^2) A^6}{720} \Bigg) \Bigg)
\end{multline}
where:
\begin{itemize}
    \item $a$: Earth's equatorial radius
    \item $e$: Eccentricity of the Earth
    \item $e'$: Second eccentricity
    \item $\phi$: Latitude
    \item $k_0$: Scale factor
    \item $\nu$: Radius of curvature in the prime vertical
    \item $A$: Difference in longitude from the central meridian
    \item $T$: Tangent of latitude
    \item $C$: Meridian convergence
    \item $x_0$: False easting
    \item $y_0$: False northing
\end{itemize}
The previous formulas are complex for the given task, so an approximation described \cite{mapping} was used:

% Latitude and Longitude to Radians
\begin{equation}
\text{latitude}_{\text{rad}} = \frac{\text{latitude}_{\text{deg}} \times \pi}{180}
\end{equation}

\begin{equation}
\text{longitude}_{\text{rad}} = \frac{\text{longitude}_{\text{deg}} \times \pi}{180}
\end{equation}

% Conversion to x and y
\begin{equation}
x = 6378137 \times \text{longitude}_{\text{rad}} + 500000
\end{equation}

\begin{equation}
y = 6378137 \times \ln \left( \tan \left( \frac{\pi}{4} + \frac{\text{latitude}_{\text{rad}}}{2} \right) \right)
\end{equation}

% \begin{gather}
% \text{latitude}_{\text{rad}} = \frac{\text{latitude}_{\text{deg}} \times \pi}{180}
% \\
% \text{longitude}_{\text{rad}} = \frac{\text{longitude}_{\text{deg}} \times \pi}{180}
% \\
% % Conversion to x and y
% x = 6378137 \times \text{longitude}_{\text{rad}} + 500000
% \\
% y = 6378137 \times \ln \left( \tan \left( \frac{\pi}{4} + \frac{\text{latitude}_{\text{rad}}}{2} \right) \right)
% \end{gather}

While the formulae above provide a general perspective, it's important to realize their potential variations based on the chosen map projection and geographic region. In our methodology, to improve scalability and reduce training times, an offset is deducted from the coordinates. This ensures that each training sequence commences at the coordinate origin (0, 0).

The process of data collection entails live streaming data into Rosbag files \cite{rosbag} as the vehicle traverses the intended route. The creation of these Rosbags is facilitated by the Aslan project \cite{aslan}, an open-source venture built atop Robot Operating System (ROS). Each Rosbag captures a distinct route, with the offset already subtracted. Since these sensors operate at varying frequencies and to maintain data integrity for future algorithmic processes, it's important to interpolate the data, ensuring uniform timestamps across all sensor readings.

\subsection{Polynomial Regression with C++ and L-BFGS}

Polynomial Regression is used for two steps: interpolation of the input data and sampling of the vehicle's current position at the end.
\begin{enumerate} % Numbered list
    \item Interpolations at the beginning are used over the input data to match the sampling frequencies of the different features. The position has noticeably lower frequency than IMU data. Interpolation will create the sequences of samples with the same time stamps and frequency.
    
    \item Sampling at the end is used to find the car's exact position in real-time. The Polynomial Regression fits the curve through future prediction and previous samples. It allows the sampling of the current position between the last sample and the future one.
\end{enumerate}

L-BFGS \cite{wang2015} is often favored over Gradient Descent due to its approximation of the objective function's second derivative. This approximation provides a more refined search direction. While Gradient Descent might require more iterations to achieve convergence, it is typically more straightforward to implement.

Given the inefficiencies of the Python programming language and the SciPy library for this specific problem, transitioning to C++ became essential. C++ offers notable performance enhancements, particularly for problems of this nature, and when combined with the GSL-GNU Scientific Library \cite{gsl}, it yields exceptional results. The GNU Scientific Library (GSL) is a comprehensive numerical library tailored for C and C++ developers. Freely available under the GNU General Public License, this library equips programmers with a lot of mathematical routines, including but not limited to, random number generators, special functions, and least-squares fitting.

In the C++ algorithm, L-BFGS optimization is employed alongside third-order polynomial regression. The decision to opt for the third order is informed by an analysis of the collected data. Metrics such as GPS coordinates, acceleration, and orientation typically manifest as simple functions—often linear. However, in more complex scenarios, like when a vehicle takes turns, these functions can shift to second or even third order representations. There are no situations where a higher order of polynomial regression would be needed because of the vehicle speed constraints and inertia. The regression model can be denoted as:
\begin{equation}
\textit{Position} = \theta_0 + \theta_1 \cdot t + \theta_2 \cdot t^2 + \theta_3 \cdot t^3
\end{equation}

\subsection{Data Preparation}

The subsequent phase involves processing the data to extract as much useful information as possible. The data should be segmented into sequences to fully harness the capabilities of Recurrent or LSTM Neural Networks. Typically, time steps between successive samples from GPS sensors range between 100 ms and 600 ms, indicating a variable sampling frequency. Such fluctuations are attributed to factors like changing weather conditions, terrain, infrastructure of the area, tall obstacles, and sometimes proximity to power grids. These inconsistencies pose challenges for autonomous vehicles that rely on real-time location data. The proposed solution involves using the algorithm to forecast a new location coordinate 200 ms into the future. This 200 ms value was chosen based on the average time step observed between samples acquired via a GPS sensor over several recorded routes (Rosbags). By forecasting the vehicle's future location, we can compute the vehicle's precise position using a predetermined interpolation algorithm that takes into account both the predicted and prior locations. The subsequent graphs illustrate the inconsistency in coordinate frequency. To obtain synchronous data samples at a fixed frequency, we interpolated both the IMU readouts and the GPS observations and evaluated the regression models at regular 200 ms intervals.

\begin{figure}%[h]
    \centering
    \includegraphics[width=\linewidth]{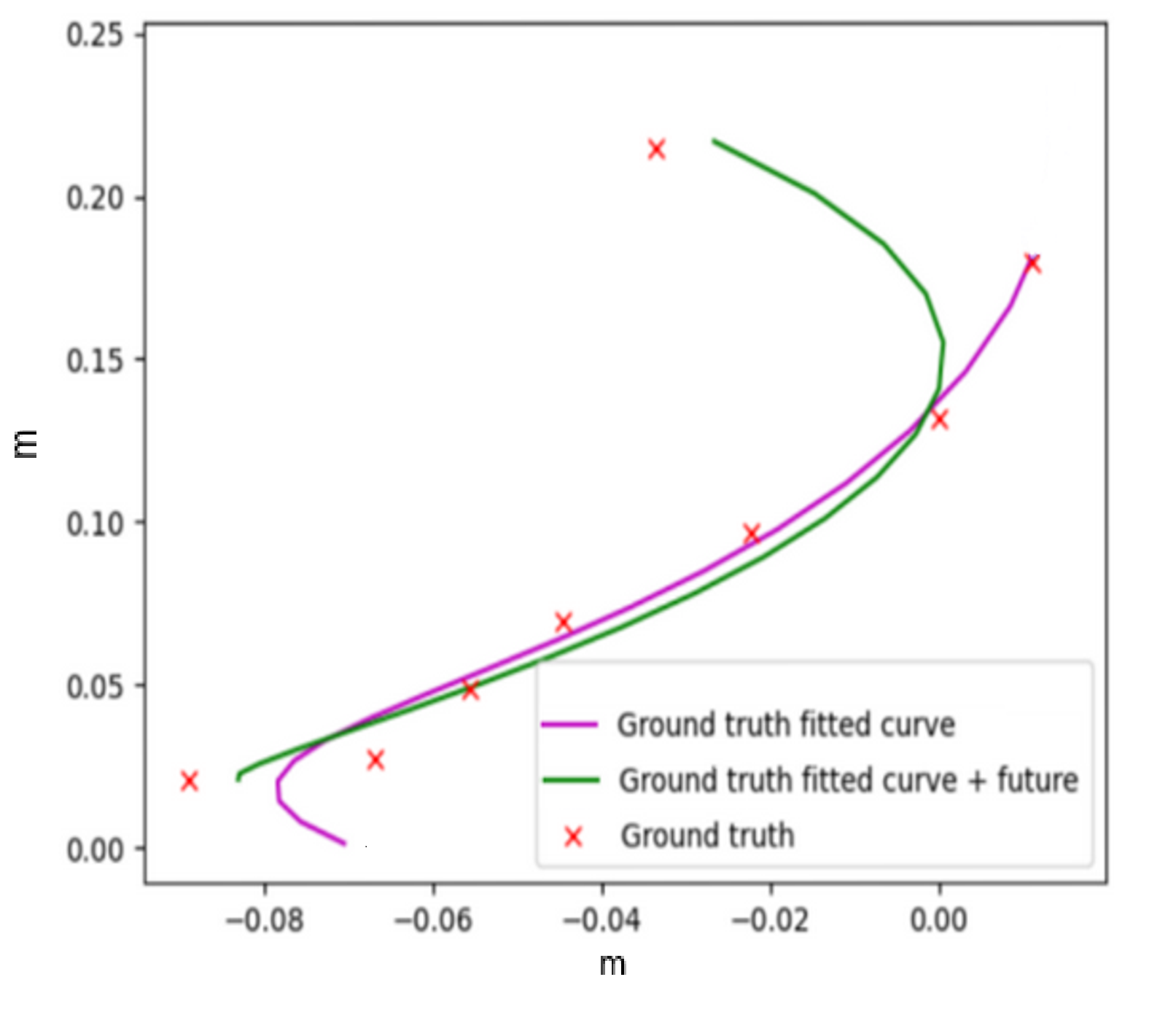}
    \caption{Visualisation of fitted curve relation with given coordinates while taking a turn.}
    \label{fig:curve_turn}
\end{figure}

Figure \ref{fig:curve_turn} depict raw predictions derived directly from the regression model, trained on known coordinates (excluding future coordinate), with time intervals of 200 ms from the preceding known coordinate. These graphs reveal how a single future point can significantly alter a car's orientation during a turn, which might result in unintended consequences. To ensure the data is well-suited for the neural network, we interpolate the data to achieve a consistent time step/frequency of 200 ms. This task is executed using Polynomial Regression, as detailed in section Data Preparation. This uniformity is vital since variable sampling rates can complicate LSTM Neural Network's task. The objective is to maintain high model accuracy while simplifying its structure, thus reducing its parameters. The preprocessing steps are as follows:

\begin{enumerate}
    \item Loading saved files from various routes.
    \item Segmenting the GPS data into 10-sample sequences with a stride of one.
    \item Constructing functions by fitting the GPS coordinate sequences using Polynomial Regression.
    \item We aim to generate eight interpolated GPS coordinate samples, each spaced by 200 ms, resulting in a total span of 1.6 seconds. To accomplish this, an interpolator is constructed using a sufficient number of the most recent GPS measurements, ensuring that the range covers at least a 2-second interval. This approach allows the set of eight targeted samples to be accurately interpolated within the collected input points, ensuring a reliable representation of the positional changes over the specified duration. By doing this, we ensure that the interpolator is constructed with an adequately rich set of points, allowing for the generation of the eight desired samples with precise 200 ms intervals.
    \item Synchronizing acceleration and orientation data with GPS coordinate timestamps. Notably, the IMU sensor's frequency is approximately five times that of the GPS sensor.
    \item Annotating the data as either straight-path or containing a turn, as described bellow.
\end{enumerate}

This final preprocessing step (Annotation) is important for training the neural network. A considerable portion of the data represents a linear path with consistent orientation and acceleration. Only a minuscule fraction of the turns can be discerned through the accelerometer by monitoring acceleration and deceleration along the Y-axis. To address this imbalance, the data will be annotated to prioritize sampling in the data loader or to be augmented/duplicated. This ensures the model predicts with comparable accuracy during turns as it does on straight paths.

\textbf{Important notes:} 
\begin{itemize}
    \item The initial N samples are disregarded owing to the high noise levels observed when the sensors are started. This is to avoid inadvertently training the model to predict noise. The same logic applies when the vehicle halts at stop signs or traffic lights, as these scenarios utilize a different prediction algorithm.
    \item To effectively utilize Polynomial Regression for data fitting, it's crucial to eliminate the time offset, represented by the initial sample's timestamp. This adjustment is necessary to fix the problem of input and output data having different scales and to decrease training time, and it ensures that the timeline always starts at zero.
\end{itemize}

\subsection{Model Architecture}

The model is made up of an input layer Multi-Layer Perceptron (MLP), an LSTM processing unit, and a output layer MLP. Wrapping the LSTM with two MLPs can enhance accuracy by extracting patterns from features and decoding the output to its original dimensionality. 

\subsubsection{Role of MLPs}
The input MLP learns a high-level representation of the input data. Activation functions in the MLP layer form the encoded data, which can capture crucial patterns or relationships beneficial for subsequent processing tasks. On the other hand, the output MLP discerns correlations and dependencies between features, encoding them for further analysis.

\subsubsection{Role of LSTM}
LSTM layer efficiently process sequence data. They regulate the flow of information into and out of the cell, allowing them to selectively remember or forget data over extended sequences. This is particularly useful in our context for identifying influential samples in the prediction process.

\subsection{Model Details}

\subsubsection{Input Data}
The model receives data shaped as [Nb, 6, 8], where Nb represents the batch size, 6 signifies feature count, and 8 denotes sequence length. The rationale behind the batch size and shape will be elaborated later.

\subsubsection{Model Components}

The architecture encompasses three principal components:
\begin{enumerate}
    \item MLP Encoder
    \item LSTM layer
    \item MLP Decoder
\end{enumerate}
A Leaky ReLU activation function is employed. This function introduces non-linearity with a slight positive slope for negative inputs, reducing the risk of "dying neurons" and thus improving deep neural network training.

The activation function is described by: 
\[ f(x) = \max(\alpha x, x) \]
where the slope is controlled by parameter \(\alpha\), set to 0.1 in this architecture.

\begin{figure}%[ht]
    \centering
    \includegraphics[width=\linewidth]{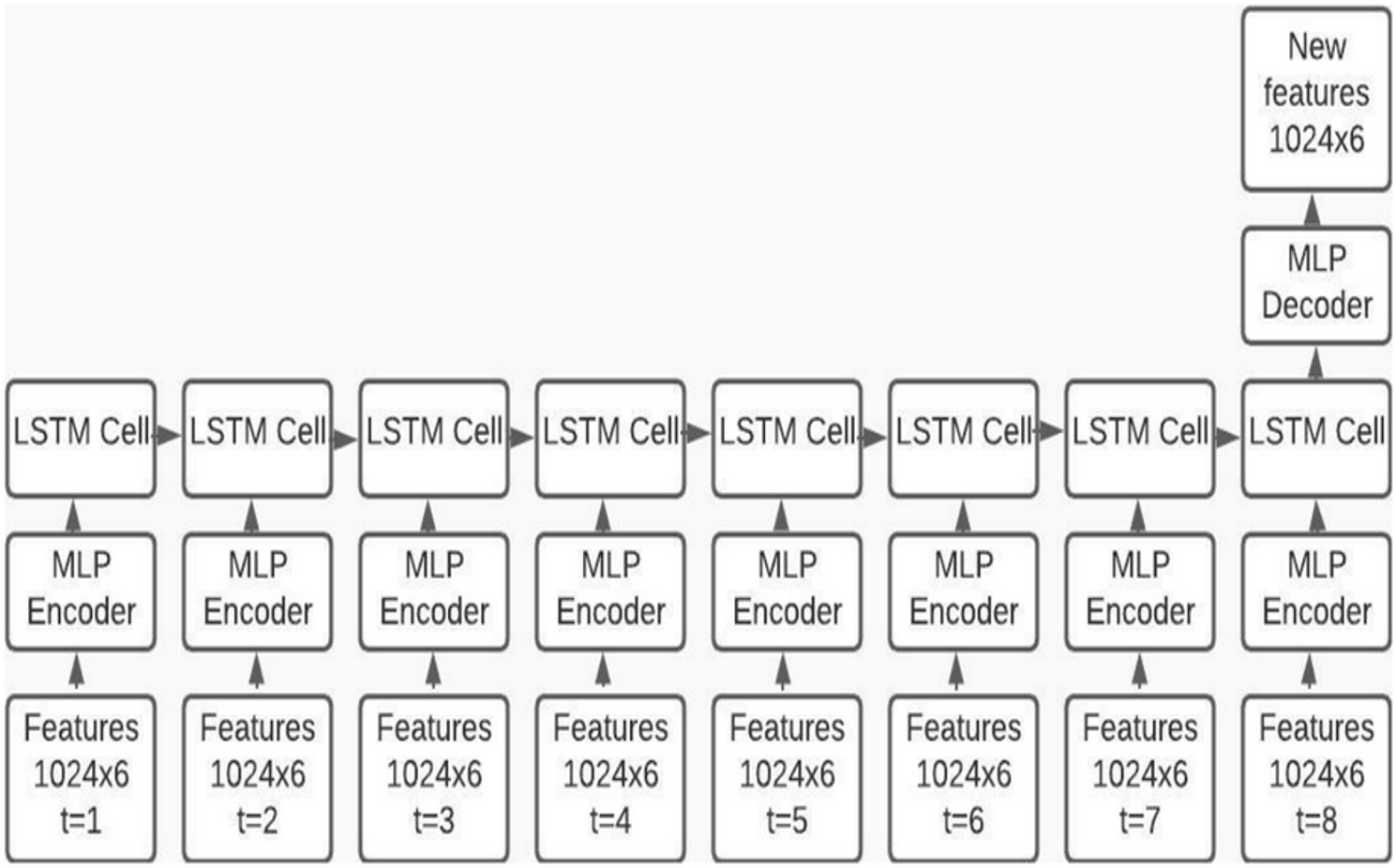}
    \caption{The block diagram of the model's architecture.}
    \label{fig:graph}
\end{figure}

\subsubsection{Visualization and Further Expansion}
Figure \ref{fig:graph} illustrates the complete runtime architecture via a block diagram. The model's architecture is based on a multiple-input, single-output configuration for runtime settings and a multiple-input, multiple-output configuration for training scenarios as described in \cite{yan2020} and \cite{jiao2016}. In the runtime, the model will give a single prediction of the future point (single-output), which the interpolator will use to sample the current position. In training, the model will produce eight predictions; the first seven predictions should represent the already known points fed into the model (the first prediction is equivalent to the second input sample, the second prediction is equivalent to the third input sample, etc.), and the last eighth point, the future point. By including these additional points next to future points, training time is significantly reduced, and accuracy is improved. The numbers $1024$ and $6$ respectively indicate the batch size and the feature count, including $x$ and $y$ coordinates, acceleration measurement $a_x$ and $a_y$, vehicle orientation $\theta_{\text{yaw}}$, and an optional timestamp $t$.

It's possible to augment the model by incorporating an additional LSTM layer, providing:
\begin{itemize}
    \item Enhanced model capacity
    \item Superior representation learning
    \item Improved generalization
    \item Ensemble learning benefits
\end{itemize}

The model was initially trained on a GPU utilizing CUDA to optimize speed and performance. However, given the model's lightweight architecture, it operates efficiently on a CPU during runtime.

\subsection{Dataloader and Training}

\subsubsection{Data Structure}

To initiate model training, a dataloader must be established. This dataloader generates mini-batches of data in the format [Nb, 6, 8]. The number 6 represents the following six features:

\begin{enumerate}
    \item \(\Delta X\) GPS coordinate
    \item \(\Delta Y\) GPS coordinate
    \item Acceleration on X-axis
    \item Acceleration on Y-axis
    \item Yaw angle
    \item Timestamp of the sample (optional)
\end{enumerate}

The number 8 signifies the count of samples in each sequence. Timestamps originate from the initial sequence sample, with the inaugural sample timestamped at 0 and subsequent samples increasing in 200 ms increments. Including the Timestamp feature won't increase the final accuracy as it has fixed values, but it appears useful for troubleshooting purposes and faster training.

\subsubsection{Feature Adjustments}

Rather than directly employing raw GPS coordinates, the model utilizes \(\Delta X\) and \(\Delta Y\) to represent the traveled distance. This measures the differential position between the current and preceding sample. By initializing the first sample to (0, 0) m, we enhance the model's performance and decrease training time by eliminating any substantial position offsets.
Specifically, \(\Delta X\) and \(\Delta Y\) are computed by subtracting GPS coordinates of successive samples, commencing from the future data point—our primary training ground truth—and progressing to the first sequence sample, which is set to (0, 0).

\subsubsection{Timestamp Role}

While the inclusion of timestamps might not significantly affect the model's precision or training duration, they offer valuable insights during system debugging and intuitive model evaluation.

\subsubsection{Acceleration Coordinate Transformation}

Acceleration metrics are gathered in a vehicle-relative coordinate system, necessitating a conversion to a global coordinate system. Though this transformation might not markedly amplify model accuracy, it can potentially expedite the training process. The transition between coordinate systems is described by:

\begin{equation}\label{eq:rotation_matrix}
R = \begin{bmatrix}
    \cos \theta & -\sin \theta \\
    \sin \theta & \cos \theta \\
\end{bmatrix}
\end{equation}

\begin{equation}\label{eq:transformation}
A_G = A_vR
\end{equation}

Equations \eqref{eq:rotation_matrix} and \eqref{eq:transformation} describe the computation of the rotation matrix and the subsequent acceleration transformation into global coordinates.

\subsection{Iteration}

\begin{enumerate}
    \item \textbf{Feedforward} \\
    Dataloader will pass the batch to the model. The model will perform standard feedforward with standardization, except for creating the output. \\
    The model takes a sequence of eight samples spaced each 200 ms in between. It will create a prediction for each sample 200 ms after the sample's timestamp. In this logic, providing the model with the last sample (the newest measurement) will give the prediction in the future. The model creates eight predictions from eight input samples, with the last one as a future prediction and the previous seven corresponding to input samples. As previously mentioned, the model will return either the last prediction (future) or all eight predictions, depending on the runtime or training configuration.

    \item \textbf{Backpropagation} \\
    After generating the eight predictions (the last prediction is the prediction in the future) model calculates the two losses which will be combined into one in the following manner:
    \begin{enumerate}
        \item Calculates the first loss between all outputs from the model (eight outputs) ground truth samples. The model is fed with eight samples, and it outputs eight predictions. The first seven predictions correspond to 7 fed samples which will now act as a ground truth. The last prediction corresponds to the original future sample ground truth. This loss will evaluate all predictions from the model.
        \item Calculates the second loss between the ground truth future sample and last predicted sample only on GPS (X, Y) coordinate features.
        \item Summing the loss with the following equation: 
        \[
        \text{loss} = \text{loss}_1 + \text{loss}_2 \times 10
        \]
        This prioritises the GPS coordinates because the GPS coordinates are the main feature that will be used and the IMU sensor has a much greater sampling frequency.
    \end{enumerate}
    Backward propagates with the obtained loss. If the GPS sensor falls into the delayed state, this process can be repeated by using the predicted point with previous points to predict another point.
\end{enumerate}

To combat slow training and overfitting, we introduce the learning rate scheduler (ReduceLROnPlateau).

\subsection{Handling Dropped GPS Samples}
In the runtime configuration, if a GPS sample is dropped, the system will adapt by taking the previous prediction from the model as the last sample of the new input sequence, maintaining equal spacing (as a new measurement would have). While the acceleration, speed and orientation are usually taken from previous predictions, in this case, they will be directly sampled from the IMU sensor due to its accuracy and high frequency which allows that, ensuring maximum performance. Although predictions for these features are primarily used for training purposes and gives reasonably accurate results, relying on actual IMU data proves to be more effective for optimum system performance.

\subsection{ONNX optimization}
Utilizing the ONNX (Open Neural Network Exchange) \cite{onnx} makes neural network models work more smoothly and efficiently across various AI frameworks and hardware platforms. By converting our model from PyTorch to ONNX, we have achieved a 30\% reduction in latency, enhancing the model’s processing speed

\section{Experiments and Results}

\subsection{Handling Vehicle Stops}

Data from the IMU sensor is filtered through a Low-Pass filter to eliminate high-frequency noise, smoothing the speed and acceleration readings. However, due to the inherent drift and bias in the sensor, the data tends to be unreliable when the vehicle is stationary. To address this problem, we use speed readings from the vehicle's CAN bus, which offers more reliable measurements. Given that no sensor is 100\% accurate, a threshold of 0.15 km/h has been used; speeds below this threshold indicate that the vehicle is stationary. Polynomial Regression is unnecessary because the position can be used from previous samples with "stationary" speed (speed below threshold). The second reason for not using the previously mentioned Regression for interpolation is in case of a long stop. Regression would produce noise/outliers as it cannot fit a curve through static points.

We explored the outcomes of the proposed algorithm, adapted for a Python environment. The results, detailed in graphs and tables, span three sub-sections:
\begin{itemize}
    \item Evaluation of LSTM model accuracy.
    \item Comparison to conventional methods.
    \item Inference time.
\end{itemize}

\subsection{Evaluation of LSTM Model Accuracy}

Post-training, the model predicts 200 ms ahead. The vehicle's current position is obtained by fitting the curve between the present sequence and the following future prediction and then sampling it with the current timestamp, enhancing robustness against outliers frequent in GPS readings.

\begin{figure}
\centering
\includegraphics[width=\linewidth]{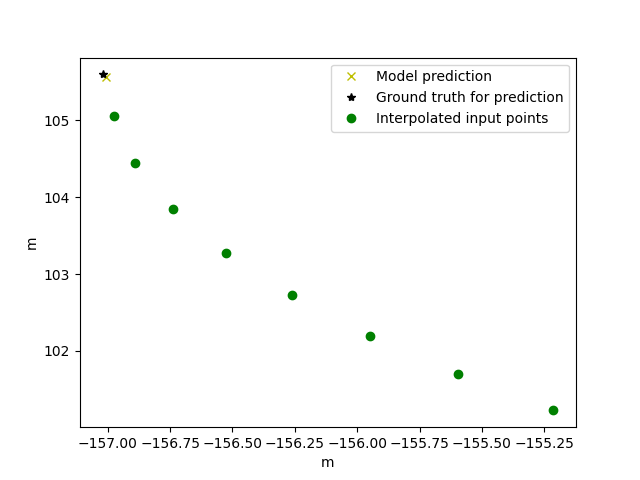} % Change "path_to_your_image" to the actual path of your image
\caption{Model prediction, input position sequence, and future point ground truth visualization.}
\label{fig:model_pred}
\end{figure}

Figure~\ref{fig:model_pred} exhibits the model output and part of the input with ground truth for the prediction. Queued interpolated positions (position features of input data) are shown as green dots. The yellow cross displays the Model prediction, and the black star shows the ground truth for that prediction (in the future).

\begin{figure}
\centering
\includegraphics[width=\linewidth]{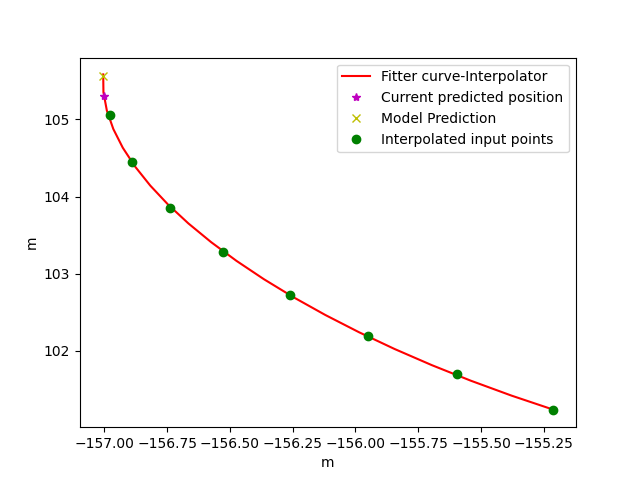}
\caption{Current (real-time) position prediction.}
\label{fig:truth_interp}
\end{figure}

Figure~\ref{fig:truth_interp} visualizes the final result of the system. This figure displays the vehicle's current position between the future prediction and the last measurement. This position is estimated with Polynomial Regression as previously described.
The red line shows a fitted curve through interpolated points from previous GPS measurements. Queued interpolated positions (position features of input data) are shown as green dots. The yellow cross displays the model prediction, and the magenta star shows the final output of the system, which represents the vehicle's current position.

The dataset comprises 19 pickle files extracted from Rosbags, representing different routes. After segregating them into training and validation sets, the LSTM achieved an average accuracy of 11 cm on the validation set.

\begin{table}
\centering
\caption{Average error on various route types.}
\begin{tabular}{|c|c|}
\hline
Route type & Average error [cm] \\
\hline
Straight path / Full speed & 16.6 \\
Turn & 12.1 \\
Roundabout & 9.7 \\
Starting / Stopping & 4.2 \\
\hline
\end{tabular}
\label{tab:avg_error}
\end{table}

\begin{figure}
\centering
\includegraphics[width=\linewidth]{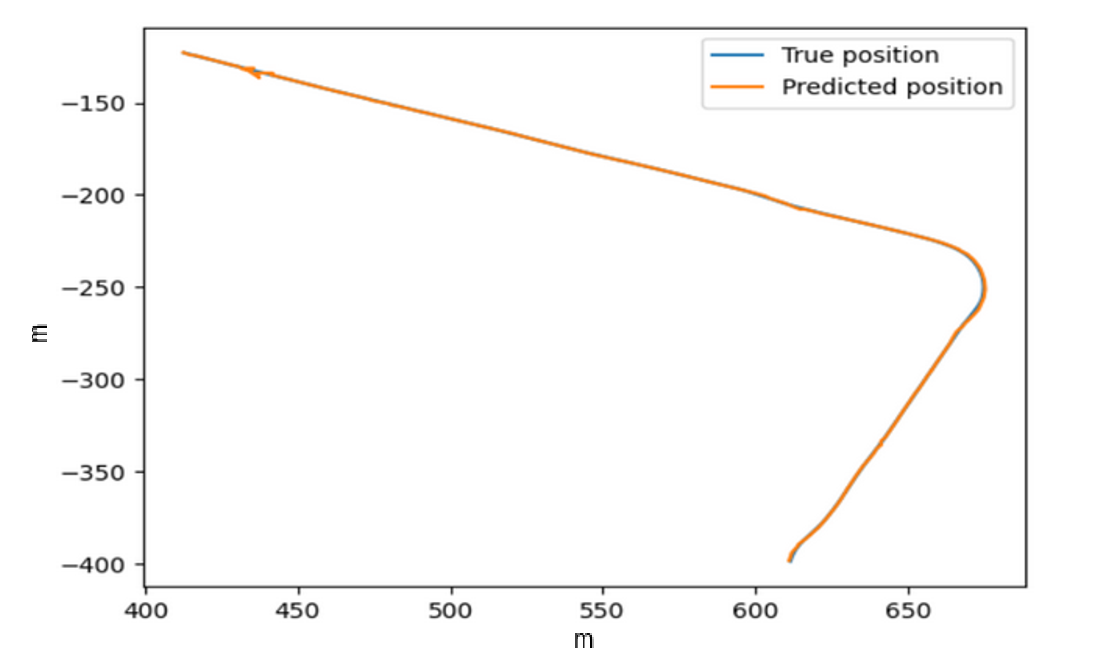}
\caption{Comparison of predicted path to the ground truth.}
\label{fig:path_compare}
\end{figure}

Table~\ref{tab:avg_error} exhibits the error trends. Notably, as the vehicle's speed rises, the prediction error increases proportionally.

Finally, Figure~\ref{fig:path_compare} contrasts the predicted path against the actual one, verifying the algorithm's efficacy.

\subsection{Comparison to Conventional Methods}

A common and straightforward method to predict an object's future location based on GPS and IMU sensor readings is the Kalman Filter. This filter operates without training compared to the LSTM Neural Network. Utilizing the known eight interpolated samples, it computes parameters—primarily the Kalman Gain, covariance matrices—and subsequently updates the system state to predict one sample into the future.

\begin{figure}
\centering
\includegraphics[width=\linewidth]{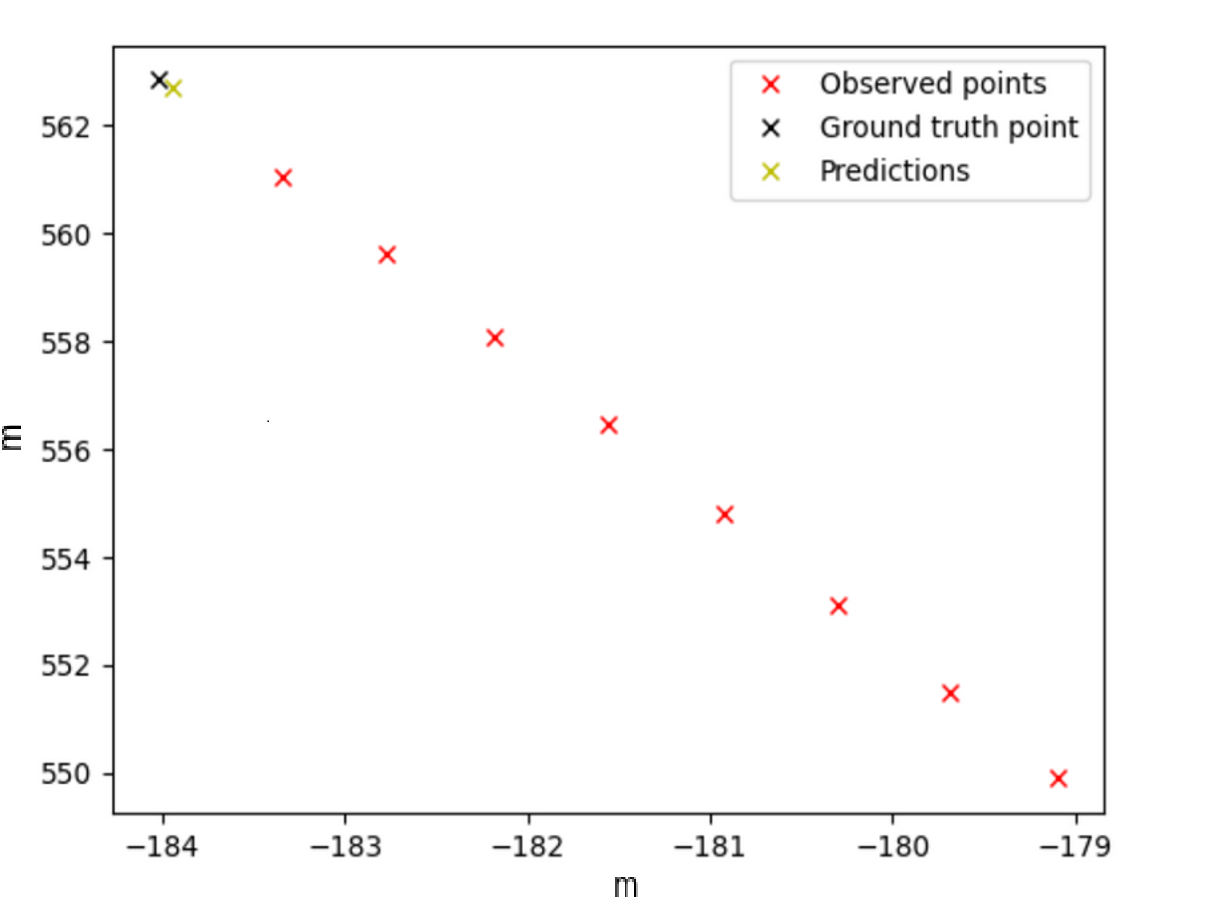}
\caption{Kalman Filter prediction visualization.}
\label{fig:kalman_pred}
\end{figure}

The results from the Kalman Filter are depicted in Figure~\ref{fig:kalman_pred}. The red crosses correspond to the known points, the black cross denotes the ground truth, and the predicted point is showcased as a yellow cross. Notably, the average prediction error on the validation dataset using the Kalman Filter is 46 cm.

\subsection{Inference Time}

The LSTM network, as discussed, is versatile, with the capability to run on both GPU and CPU hardware. Given that the data processed is relatively lightweight in terms of memory consumption, either hardware choice can yield results. The GPU is used for the training configuration as it leads to shorter training time, but in the runtime configuration, the CPU is used as it proved to have similar (or even shorter) inference time because of the model and input data simplicity and size.

To provide an aggregate estimation of the inference time, the computational duration of each component of the algorithm is summed up:
\begin{itemize}
\item Inference time for Polynomial Regression: $\sim$ 2 ms
\item Inference time for the LSTM Neural Network: $\sim$ 2 ms
\item Inference time for the Kalman Filter: $\sim$ 1 ms
\end{itemize}

The algorithm leveraging the LSTM network operates in a step-wise manner:
\begin{enumerate}
\item Polynomial Regression is employed to fit the data onto fixed time intervals.
\item The model is utilized to predict a new sample projected 200 ms into the future.
\item Interpolation is performed to ascertain the position for a specific interval timestamp between the current time and 200 ms in the future.
\end{enumerate}

The total computational time for these steps amounts to approximately 6 ms. However, it is essential to note that the first step is only occasionally executed, specifically when we need to predict more than 200 ms into the future. For autonomous driving scenarios, where pinpointing the vehicle's exact position every 50 ms is crucial, the first iteration would take roughly 6 ms. In contrast, the subsequent three iterations would require about 4 ms. Thus, depending on the specific use case, the inference time can range between 4 ms to 6 ms.

\section{Discussion}

The presented methodology, combining LSTM networks and polynomial regression, offers a robust and effective way to predict GPS coordinates in real time. Its applicability and results bear testament to the immense potential of hybrid approaches in addressing challenges that have conventionally been tackled by linear methods or filter-based systems, such as the Kalman filter.

LSTMs inherently capture temporal dependencies in sequences, making them prime candidates for time-series forecasting tasks. With polynomial regression, we are effectively addressing the issues of variable sampling rates and providing a continuous representation of the data. This fusion ensures the algorithm's resilience, even in circumstances with inconsistent data streams or sparse data.

However, as outlined in the results, it is evident that the approach does have its drawbacks, notably the increase in prediction error as vehicle speed rises. This trend suggests that the LSTM may not be predicting the dynamic changes in movement as efficiently as desired, especially in high-speed scenarios.

One of the causes is simple physics; as the vehicle moves at a higher speed, it travels a greater distance, so the 1\% of the error is a more considerable distance than when the car is moving slowly or taking a turn.

Moreover, when comparing the LSTM approach to the conventional Kalman filter, the LSTM clearly exhibits superior accuracy. This signifies the potency of deep learning methodologies in tasks that previously relied on traditional filtering techniques. That said, the Kalman filter still holds its merits, especially in its simplicity and computational efficiency.

\section{Conclusion and Future Work}

From the results presented in the preceding section, we can conclude that the proposed system achieves its objectives successfully. Despite the challenges highlighted, the algorithm delivers on its purpose, with certain limitations. 

The heaviest requirement is to train the algorithm on a diverse range of tracks or rosbags. This approach would deter the problem of overfitting and, in turn, enhance accuracy on previously unknown tracks. The aim is to provide a precise future location prediction, especially when there are inconsistencies of noisy measurements. The variable sampling rate introduced by GPS sensors is addressed by creating a ground truth derived from the fitted curve, which yields superior results, particularly in scenarios where GPS readings may be dropped or outliers emerge.

The LSTM model exhibits commendable performance, ensuring a degree of accuracy suitable for vehicle location predictions. However, it may give worse results when exposed to data that deviates significantly from the training set. Even in such situations, it offers a degree of precision comparable to the Kalman Filter, all the while ensuring sufficiently short inference time. Notably, the lightweight nature of the algorithm permits it to reside within RAM or GPU without hindering other concurrent processes.

While the development and implementation of this system meet the requirements for the scenarios discussed, precise operations may mandate the following future enhancements:

\begin{itemize}
    \item Comparative studies against the Extended Kalman Filter, which is tailored for nonlinear challenges.
    \item The potential integration of alternative decoders or encoders. For instance, the introduction of self-attention modules \cite{xie2019} could be instrumental in consolidating vital data.
    \item Consideration of ensemble methods which amalgamate the capabilities of the Extended Kalman Filter, the LSTM model, and other potential models to optimize prediction accuracy.
\end{itemize}

\bibliographystyle{IEEEtran}

\begin{thebibliography}{99}

\bibitem{liu2023}
Sheng Liu, Vivekanandh Elangovan, Weidong Xiang.
\newblock A Vehicular GPS Error Prediction Model Based on Data Smoothing Preprocessed LSTM.
\newblock \textit{Electrical and Computer Engineering, University of Michigan-Dearborn, Dearborn MI, USA}, 2023.

\bibitem{yang2020}
Sun Yang, Peng Xinya, Ding Zexuan, Zhao Jiansen.
\newblock An Approach to Ship Behavior Prediction Based on AIS and RNN Optimization Model.
\newblock \textit{Merchant Marine College, Shanghai Maritime University, Shanghai, China}, 2020.

\bibitem{altche2023}
Florent Altché, Arnaud de La Fortelle.
\newblock An LSTM Network for Highway Trajectory Prediction.
\newblock 2023.

\bibitem{gao2018}
Miao Gao, Guoyou Shi, Shuang Li.
\newblock Online Prediction of Ship Behavior with Automatic Identification System Sensor Data Using Bidirectional Long Short-Term Memory Recurrent Neural Network.
\newblock \textit{Navigation College, Dalian Maritime University, Dalian, China}, 2018.

\bibitem{alam2020}
Omar Alam, Anshuman Kush, Ali Emami, Parisa Pouladzadeh.
\newblock Predicting irregularities in arrival times for transit buses with recurrent neural networks using GPS coordinates and weather data.
\newblock 2020.

\bibitem{wang2015}
Wang, Yue-Peng \& Tao, Sulin \& Chen, Qun.
\newblock Retrieving the variable coefficient for a nonlinear convection–diffusion problem with spectral conjugate gradient method.
\newblock \textit{Inverse Problems in Science and Engineering}, 23: 1-24, 2015.

\bibitem{feng2017}
Feng, Weijiang \& Guan, Naiyang \& Li, Yuan \& Zhang, Xiang \& Luo, Zhigang.
\newblock Audio visual speech recognition with multimodal recurrent neural networks.
\newblock 681-688, 2017.

\bibitem{jenkins2018}
Jenkins, Ian \& Gee, Ludvig \& Knauss, Alessia \& Yin, Hang \& Schroeder, Jan.
\newblock Accident Scenario Generation with Recurrent Neural Networks.
\newblock 3340-3345, 2018.

\bibitem{yan2020}
Yan, Sibo \& Gu, Yuechun.
\newblock Price Forecast in High-Frequency Stock Market: An Autoregressive Recurrent Neural Network Model with Technical Indicators.
\newblock 2020.

\bibitem{jiao2016}
Jiao, Yishan \& Tu, Ming \& Berisha, Visar \& Liss, Julie.
\newblock Accent Identification by Combining Deep Neural Networks and Recurrent Neural Networks Trained on Long and Short Term Features.
\newblock 2388-2392, 2016.

\bibitem{xie2019}
Xie, Jun \& Chen, Bo \& Gu, Xinglong \& Liang, Fengmei \& Xu, Xinying.
\newblock Self-Attention-Based BiLSTM Model for Short Text Fine-grained Sentiment Classification.
\newblock \textit{IEEE Access}, 7: 1-1, 2019.

\bibitem{cremanns2017}
Cremanns, Kevin \& Roos, Dirk.
\newblock Deep Gaussian Covariance Network.
\newblock 2017.

\bibitem{mapping}
\newblock John P. Snyder
Map Projections - A Working Manual, by, U.S. Geological Survey Professional Paper 1395, United States Government Printing Office, Washington D.C
\newblock 1987.

\bibitem{rosbag}
ROS Wiki,
\emph{Bags}, 
Available: \url{http://wiki.ros.org/Bags}.

\bibitem{aslan}
The ASLAN Project,
\emph{The ASLAN Project: Open Source Autonomous Driving}, 
Available: \url{https://www.aslanproject.com/}.

\bibitem{gsl}
GNU Project,
\emph{GNU Scientific Library}, 
Available: \url{https://www.gnu.org/software/gsl/}.

\bibitem{onnx}
ONNX,
\emph{Open Neural Network Exchange},
Available: \url{https://onnx.ai/}.

\end{thebibliography}

\end{document}